\documentclass[sigconf]{acmart}

\usepackage{multirow}

\AtBeginDocument{%
  }

\copyrightyear{2023}
\acmYear{2023}
\setcopyright{acmlicensed}
\acmConference[MM '23] {Proceedings of the 31st ACM International Conference on Multimedia}{October 29--November 3, 2023}{Ottawa, ON, Canada.}
\acmBooktitle{Proceedings of the 31st ACM International Conference on Multimedia (MM '23), October 29--November 3, 2023, Ottawa, ON, Canada}
\acmPrice{15.00}
\acmISBN{979-8-4007-0108-5/23/10}
\acmDOI{10.1145/3581783.3611819}

\begin{document}

\title{TSSAT: Two-Stage Statistics-Aware Transformation for Artistic Style Transfer}


\author{Haibo Chen}
\authornote{Corresponding author.}
\affiliation{%
  \institution{Nanjing University of Science and Technology}
  \city{Nanjing}
  \country{China}
}
\email{hbchen@njust.edu.cn}

\author{Lei Zhao}
\affiliation{%
  \institution{Zhejiang University}
  \city{Hangzhou}
  \country{China}
}
\email{cszhl@zju.edu.cn}

\author{Jun Li}
\affiliation{%
 \institution{Nanjing University of Science and Technology}
 \city{Nanjing}
 \country{China}
}
\email{junli@njust.edu.cn}

\author{Jian Yang}
\affiliation{%
	\institution{Nanjing University of Science and Technology}
	\city{Nanjing}
	\country{China}
}
\email{csjyang@njust.edu.cn}

\renewcommand{\shortauthors}{Haibo Chen, Lei Zhao, Jun Li, \& Jian Yang}

\begin{abstract}
  Artistic style transfer aims to create new artistic images by rendering a given photograph with the target artistic style. Existing methods learn styles simply based on global statistics or local patches, lacking careful consideration of the drawing process in practice. Consequently, the stylization results either fail to capture abundant and diversified local style patterns, or contain undesired semantic information of the style image and deviate from the global style distribution. To address this issue, we imitate the drawing process of humans and propose a Two-Stage Statistics-Aware Transformation (TSSAT) module, which first builds the global style foundation by aligning the global statistics of content and style features and then further enriches local style details by swapping the local statistics (instead of local features) in a patch-wise manner, significantly improving the stylization effects. Moreover, to further enhance both content and style representations, we introduce two novel losses: an attention-based content loss and a patch-based style loss, where the former enables better content preservation by enforcing the semantic relation in the content image to be retained during stylization, and the latter focuses on increasing the local style similarity between the style and stylized images. Extensive qualitative and quantitative experiments verify the effectiveness of our method.
\end{abstract}




\maketitle

\section{Introduction}
\label{sec:intro}

Artistic style transfer is a powerful technique for image editing and art creation, whose key problem is how to separate and recombine the contents and styles of given images. Recently, the seminal work of Gatys \emph{et al.} \cite{gatys2016image} firstly proposed to leverage a pre-trained Deep Convolutional Neural Network (DCNN) to tackle this problem, which opens up the neural style transfer era. Since then, numerous neural style transfer methods have been developed. Among them, global statistics-based and local patch-based methods dominate the current style transfer field.

\begin{figure}[t]
	\centering
	\includegraphics[width=1.0\columnwidth]{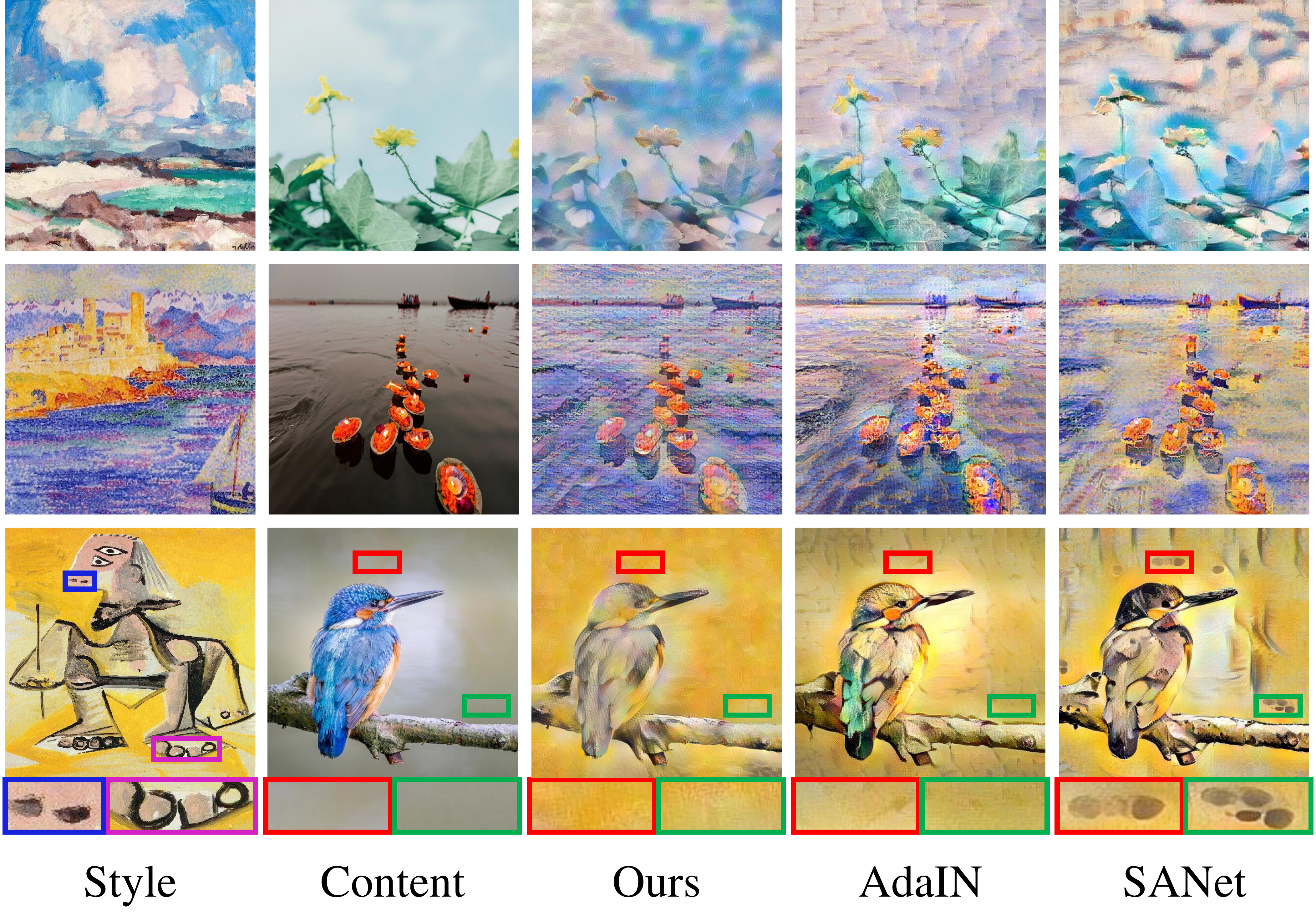}
	\caption{Stylization examples. The first two columns show the style and content images. The other three columns show the stylized images produced by our method, AdaIN \cite{huang2017arbitrary}, and SANet \cite{park2019arbitrary}.}
	\label{fig1}
\end{figure}

More specifically, global statistics-based methods \cite{gatys2016image,johnson2016perceptual,huang2017arbitrary,wang2020diversified,jing2020dynamic,lin2021drafting,hu2020aesthetic,zhang2022domain} focus on exploring proper global statistics to represent style and enforcing the global statistics of the content image to be aligned with those of the style image for stylization. For example, Huang \emph{et al.} \cite{huang2017arbitrary} found that the mean and variance of deep image features carried style information, and Li \emph{et al.} \cite{li2017universal} employed the whitening and coloring transforms to reflect the direct matching of feature covariance of the content image to a given style image. This line of work is able to align the global style distributions between the style and stylized images. However, they overlook one critical problem: a style image usually contains more than one kind of style pattern, and similarly, a content image always consists of many different semantic regions. Simply transferring the global style to the content image considers neither the diversity of local style patterns nor the difference among multiple content regions (\emph{e.g.}, AdaIN \cite{huang2017arbitrary} fails to capture the abundant style information in the $1^{st}$ row of Figure~\ref{fig1}). In contrast, local patch-based methods \cite{li2016combining,chen2016fast,sheng2018avatar,zhang2019multimodal,yao2019attention,park2019arbitrary,chen2021artistic,ijcai2022p690} conduct style transfer by replacing every content patch with similar style patches in the feature space. For example, Chen \emph{et al.} \cite{chen2016fast} performed a style-swap operation that swaps each content feature patch with its closest-matching style feature patch, and Park \emph{et al.} \cite{park2019arbitrary} introduced a style-attentional network that integrates the style feature patches according to the semantic spatial distribution of the content image. Despite the effectiveness in learning local style patterns, this line of work usually suffers from two drawbacks: 1) the style feature patches will inevitably introduce some semantic information of the style image to the stylization result (\emph{e.g.}, the stylized image generated by SANet \cite{park2019arbitrary} in the $3^{rd}$ row of Figure~\ref{fig1} contains the structure of the nose and toes of the style image); 2) some marginal colors and texture patterns in the style image may prevail in the stylization result, while those critical colors and texture patterns in the style image may be ignored. This is because these methods only consider the semantic correspondence between the content and style images, neglecting the global style distribution of the style image (\emph{e.g.}, the $2^{nd}$ row, $5^{th}$ column of Figure~\ref{fig1}).

Motivated by the observations and analyses above, we propose a Two-Stage Statistics-Aware Transformation (TSSAT) module, which \textbf{\emph{simulates the drawing process of humans}}, \emph{i.e.}, first drawing the basic and primary structures and textures from a global perspective and then further enhancing the paintings with delicate fine-grained details from a local perspective. Our TSSAT accomplishes a similar process with a global statistics alignment stage and a local statistics swap stage. In detail, TSSAT first aligns the global statistics of content and style features to learn global style, inspired by global statistics-based methods \cite{huang2017arbitrary,kalischek2021light,jing2020dynamic,liu2021structure,wu2022ccpl}. Then, for each local patch of the content feature, TSSAT finds its closest-matching style feature patch and \textbf{\emph{swaps their local statistics instead of the two local patches}}, which overcomes the inertial thinking of local patch-based methods \cite{chen2016fast,yao2019attention,park2019arbitrary,wu2021styleformer,li2022arbitrary} and revolutionizes local style learning. In this way, our method prevents the semantic information stored in local style patches from being introduced into the stylization result. Meanwhile, more abundant and fine-grained local style patterns are involved on the basis of learned global style distribution. Our TSSAT also allows flexible style pattern modulation by adjusting the patch size in the local statistics swap stage. Moreover, to further enhance both content and style representations, we introduce two novel losses: an attention-based content loss and a patch-based style loss. To be more specific, the attention-based content loss enforces the semantic relation in the content image to be retained during stylization, leading to better content preservation. And the patch-based style loss focuses on increasing the style similarity between the style and stylized images from a local perspective.

Overall, the main contributions of this paper can be summarized as follows:

\begin{itemize}
	\item We propose a TSSAT module that harnesses feature statistics to first build the global style foundation and then enrich local style details, significantly improving the stylization effects and providing fresh insight into the challenging style transfer problem.
	\item An attention-based content loss is introduced to enable better content preservation by enforcing the semantic relation in the content image to be retained during stylization.
	\item A patch-based style loss is introduced to increase the style similarity between the style and stylized images from a local perspective.
	\item Comprehensive experimental results show that our method outperforms state-of-the-art style transfer methods both qualitatively and quantitatively.
\end{itemize}

\section{Related Work}

\textbf{Global Statistics-based Style Transfer.} Global statistics-based methods generally transform the content features to match the global statistics of style features for stylization. Gatys \emph{et al.} \cite{gatys2016image} represented the style of an image with Gram matrix and constrained the Gram matrices of the style and stylized images to be consistent by iterative optimizations. Huang \emph{et al.} \cite{huang2017arbitrary} performed style transfer by adjusting the mean and variance of the content features to match those of the style features. Li \emph{et al.} \cite{li2017universal} conducted the whitening and coloring transforms (WCT) to endow the content features with the same statistical characteristics as the style features. Instead of directly using the first- or second-order statistical transformation to learn style, Li \emph{et al.} \cite{li2019learning} employed light-weighted CNNs to predict a learnable linear transformation matrix conditioned on an arbitrary pair of content and style images. Jing \emph{et al.} \cite{jing2020dynamic} extended the work of Huang \emph{et al.} \cite{huang2017arbitrary} by introducing a dynamic instance normalization (DIN) module that encodes a style image into learnable convolution parameters, upon which the content image is stylized. An \emph{et al.} \cite{an2021artflow} presented an unbiased style transfer framework that consists of reversible neural flows \cite{ho2019flow++} and an unbiased style transfer module (\emph{e.g.}, AdaIN \cite{huang2017arbitrary} or WCT \cite{li2017universal}) to address the content leak problem. Lin \emph{et al.} \cite{lin2021drafting} proposed a drafting network and a revision network to perform style transfer in a progressive procedure and relied on AdaIN \cite{huang2017arbitrary} to combine the style feature and the content feature. Recently, some methods \cite{sanakoyeu2018style,kotovenko2019content,kotovenko2019contentstyle,chen2021dualast,chen2021diverse,xu2021drb,zuo2022style} proposed to learn style from a collection of artworks rather than a single style image, vastly improving the quality of stylization results. The above methods significantly promote the development of style transfer. However, given that the style image usually contains more than one kind of style patterns and the content image always consists of multiple different semantic regions, it may be insufficient to use such global statistics to represent style.

\textbf{Local Patch-based Style Transfer.} Local patch-based methods generally swap local content patches with similar local style patches in the feature space for stylization. Chen \emph{et al.} \cite{chen2016fast} concatenated both content and style information into a single layer of the CNN, by swapping the textures of the content image with those of the style image. Sheng \emph{et al.} \cite{sheng2018avatar} proposed a patch-based feature manipulation module to transfer the content features to semantically nearest style features. Park \emph{et al.} \cite{park2019arbitrary} and Deng \emph{et al.} \cite{deng2020arbitrary} embedded a local style pattern in each position of the content features by mapping a relationship between the content and style features based on attention mechanism. Zhang \emph{et al.} \cite{zhang2019multimodal} clustered the style image features into sub-style components, which are matched with local content features under a graph cut formulation. Yao \emph{et al.} \cite{yao2019attention} employed self-attention as a residual to obtain the attention map, and then introduced multi-scale style swap and a stroke fusion strategy to adaptively integrate multiple style patterns into one stylized image. Huo \emph{et al.} \cite{huo2021manifold} proposed a manifold alignment-based style transfer framework which allows semantically similar regions between the output and style images share similar style patterns. Chen \emph{et al.} \cite{chen2021artistic} introduced an internal-external learning scheme and two contrastive losses to bridge the gap between human-created and AI-created artworks. Deng \emph{et al.} \cite{deng2022stytr2} proposed a transformer-based \cite{vaswani2017attention} style transfer framework that translates the content sequences based on the reference style sequences, leading to stylization results with well-preserved structures. Although these methods are effective in learning more local style patterns, the generated stylized images usually contain undesired semantic information of the style image and sometimes deviate from global style distribution.

\begin{figure*}[t]
	\centering
	\includegraphics[width=0.99\textwidth]{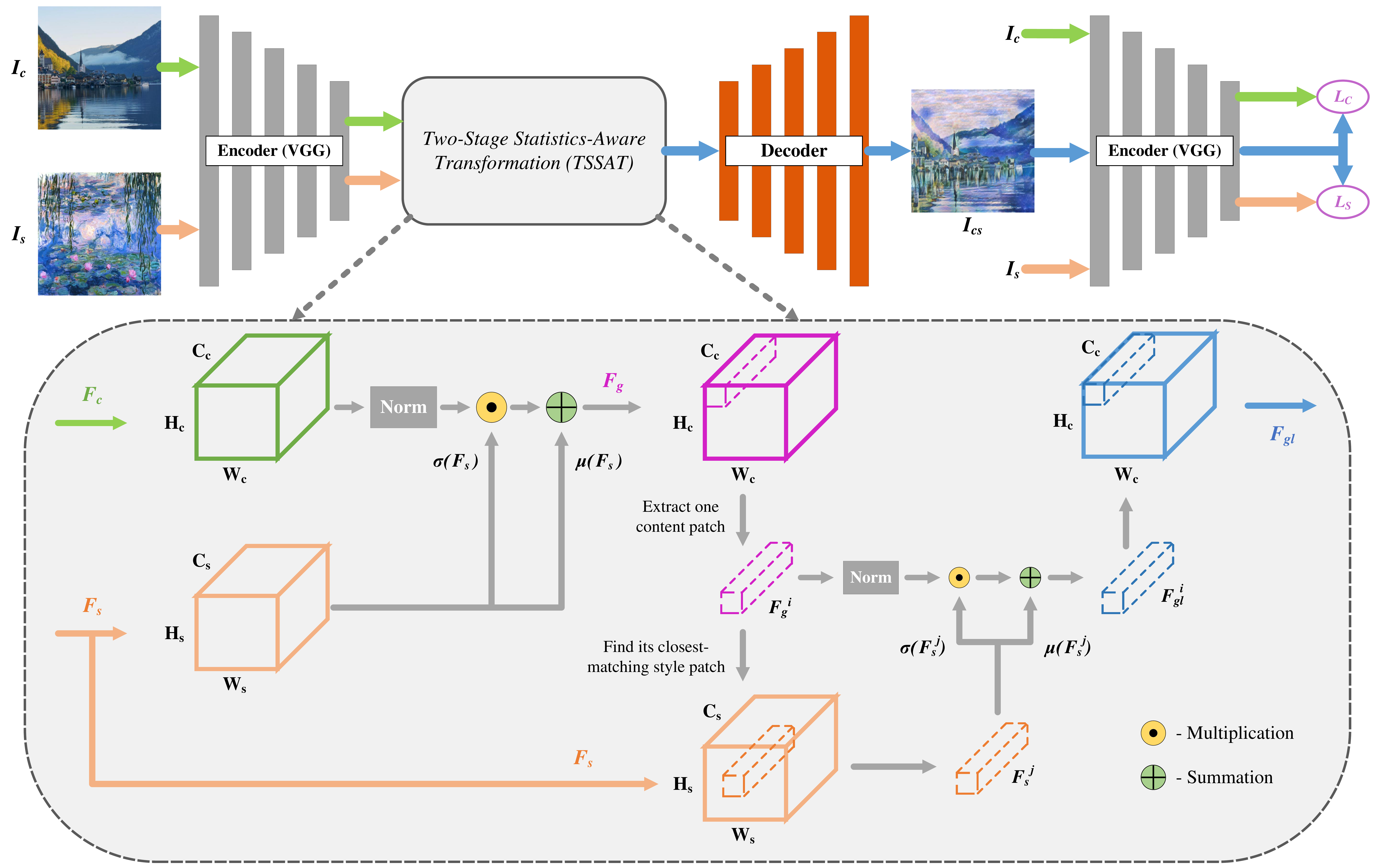}
	\caption{Overview of our framework. (1) We take a fixed pre-trained VGG-19 network as our encoder to extract content and style features. (2) The TSSAT module performs style transfer in the feature space by first aligning the global statistics of content and style features and then swapping the local statistics in a patch-wise manner. (3) The decoder inverts the deep transferred features into artistic images.}
	\label{fig2}
\end{figure*}

\textbf{Others.} Beyond that, a number of methods have been proposed to address the style transfer problem from other perspectives. Liu \emph{et al.} \cite{liu2021adaattn} integrated the ideas of Huang \emph{et al.} \cite{huang2017arbitrary} and Park \emph{et al.} \cite{park2019arbitrary} and proposed an attention and normalization module, named AdaAttN, which performs feature statistics transferring via modulation with per-point attention-weighted mean and variance of the style feature. Nevertheless, AdaAttN will inevitably lose some global style information, since it replaces the global mean and variance with the attention-weighted mean and variance that focus more on local style information. Kwon \emph{et al.} \cite{kwon2022clipstyler} introduced CLIP (Contrastive Language-Image Pre-Training) \cite{radford2021learning} into the style transfer task and used a text description instead of a style image to represent the desired style. Fu \emph{et al.} \cite{fu2022language} presented another text-guided style transfer framework that learns the correlation between text prompts and style images based on large amounts of paired training data. These methods empower users to create more creative artistic images with input texts, yet the stylization results are far from satisfactory in terms of content preservation and style transformation.

Unlike these existing methods, our approach not only takes both global and local style into consideration without involving any semantic information of the style image, but also learns better content and style representations.

\section{Proposed Method}
\label{s3}

Here, we first describe the overall pipeline of our approach in Section~\ref{s3.1}. Then, we give details of the proposed Two-Stage Statistics-Aware Transformation (TSSAT) module in Section~\ref{s3.2}. Finally, Section~\ref{s3.3} introduces the loss functions used in our model, including our newly proposed attention-based content loss and patch-based style loss.

\subsection{Overview}
\label{s3.1}

Formally, our task can be described as follows: given an arbitrary content image $I_c$ and an arbitrary style image $I_s$, we aim to learn a generative model to synthesize the corresponding stylized image $I_{cs}$ that not only preserves the content structures of $I_c$ but also learns the local and global style patterns from $I_s$. The overall framework of our approach is illustrated in Figure~\ref{fig2}. As we can see, there are mainly three components in our model: an encoder $E$, a two-stage statistics-aware transformation module $\mathit{TSSAT}$, and a decoder $D$.

In detail, the encoder $E$ is a pre-trained VGG-19 network \cite{simonyan2014very} $\phi$ whose parameters are fixed during training. We feed the content image $I_c$ and style image $I_s$ to $\phi$ to extract their respective VGG feature maps,

\begin{equation}
F_c:=\phi(I_c), \quad F_s:=\phi(I_s)
\end{equation}

After obtaining the content features $F_c$ and style features $F_s$, we employ a two-stage statistics-aware transformation module $\mathit{TSSAT}$ to match the global and local statistics of $F_c$ with those of $F_s$, yielding $F_{gl}$ as the fusion result of content and style information,

\begin{equation}
F_{gl}:=\mathit{TSSAT}(F_c, F_s)
\end{equation}

Finally, we input $F_{gl}$ to the decoder $D$ to generate the stylized image $I_{cs}$,

\begin{equation}
I_{cs}:=D(F_{gl})
\end{equation}

Besides, it is also worth noting that we use $F_{cs}$ to represent the image features extracted from $I_{cs}$ via the VGG-19 network $\phi$, $i.e.$, $F_{cs}:=\phi(I_{cs})$.

\subsection{Two-Stage Statistics-Aware Transformation}
\label{s3.2}

The key idea of TSSAT is to harness feature statistics to first build the global style foundation (in the global statistics alignment stage) and then enrich local style details (in the local statistics swap stage), simulating the drawing process of humans. In this paper, we mainly take mean and variance as the feature statistics to show TSSAT’s effectiveness. Its detailed structure is depicted in Figure~\ref{fig2} (dashed grey box).


\textbf{Global statistics alignment.} This stage aims to align the global statistics of content and style features to learn global style. Previous works \cite{huang2017arbitrary,karras2019style,lin2021drafting} have demonstrated mean and variance do not carry any semantic information but only the style information of an image. In this way, we can achieve our goal by first normalizing the content features $F_c$ and then scaling and shifting the normalized $F_c$ with the corresponding scalar components of the style features $F_s$,

\begin{equation}
F_g:=\sigma(F_s)\left(\frac{F_c-\mu(F_c)}{\sigma(F_c)}\right)+\mu(F_s)
\end{equation}
where $\sigma$ and $\mu$ denote the mean and standard deviation of feature maps, respectively. $F_g$ is the output, which captures the global style distribution of the style image.

\textbf{Local statistics swap.} In this stage, we take $F_g$ as the content features and aim to introduce more local style patterns to the stylization result by swapping the local statistics of content and style features. The detailed procedure is as follows:

\begin{enumerate}
	\item Extract a set of $k \times k$ patches for both $F_g$ and $F_s$, denoted by $\{F_{g}^{i}\}_{i \in n_g}$ and $\{F_{s}^{j}\}_{j \in n_s}$, where $n_g$ and $n_s$ are the number of extracted patches.
	\item For each content patch $F_g^i$, find its closest-matching style patch $F_s^j$ through a convolution layer, where the normalized style feature patches $\{F_{s}^{j} / \Vert F_s^j \Vert \}_{j \in n_s}$ are the filters and $F_g^i$ is the input. The output of this layer is a vector, where each scalar is equivalent to the cosine similarity between the content patch and one style patch. This way, the closest-matching style patch $F_s^j$ can be found by determine the index of the maximum value in the vector.
	\item Swap the mean and variance of the content patch $F_g^i$ with those of its closest-matching style patch $F_s^j$,
	\begin{equation}
	F_{gl}^i:=\sigma(F_s^j)\left(\frac{F_g^i-\mu(F_g^i)}{\sigma(F_g^i)}\right)+\mu(F_s^j)
	\end{equation}
	\item Recombine the feature patches $\{F_{gl}^{i}\}_{i \in n_g}$ to obtain the feature maps $F_{gl}$, where abundant and fine-grained local style patterns are involved on the basis of $F_g$.
\end{enumerate}


Note that to prevent the semantic information of the style image from being introduced into the stylization result, here we overcome the inertial thinking of previous methods \cite{chen2016fast,yao2019attention,park2019arbitrary,wu2021styleformer,li2022arbitrary} and propose to learn local styles based on feature statistics instead of neural patches, which is an elegant and effective reformation. We also emphasize that since the local statistics swap operation is performed between every two most similar patches of the global-style-aligned features $F_g$ and $F_s$, it will not result in global style deviation but only further enrich local style details (see detailed demonstrations in Section~\ref{s4.4}).

Thanks to the above two stages, our TSSAT is able to synthesize more appealing stylization results with elaborately decorated style patterns. TSSAT also allows flexible style pattern modulation by adjusting the patch size $k$ in the local statistics swap stage, leading to more diverse stylization results (see detailed demonstrations in Section~\ref{s4.2}).


\subsection{Loss Functions}
\label{s3.3}

The loss functions used in our model consist of the content loss $\mathcal{L}_c$, the attention-based content loss $\mathcal{L}_{ac}$, the style loss $\mathcal{L}_s$, the patch-based style loss $\mathcal{L}_{ps}$, and the identity loss $\mathcal{L}_{identity}$. Among them, $\mathcal{L}_{ac}$ and $\mathcal{L}_{ps}$ are our newly proposed losses. Details of each loss will be explained in the remaining part of this section.

\textbf{Content loss.} We learn the content information by minimizing the perceptual differences between the content image $I_c$ and the stylized image $I_{cs}$,

\begin{equation}
\mathcal{L}_{c}:=\sum_{i=1}^{L} \parallel \phi_{i}(I_c)-\phi_{i}(I_{cs}) \parallel_2
\end{equation}
where $\phi_{i}$ denotes the $i_{th}$ layer of VGG-19. We use relu4\_{}1 and relu5\_{}1 layers in our experiments.

\begin{figure}[t]
	\centering
	\includegraphics[width=1.0\columnwidth]{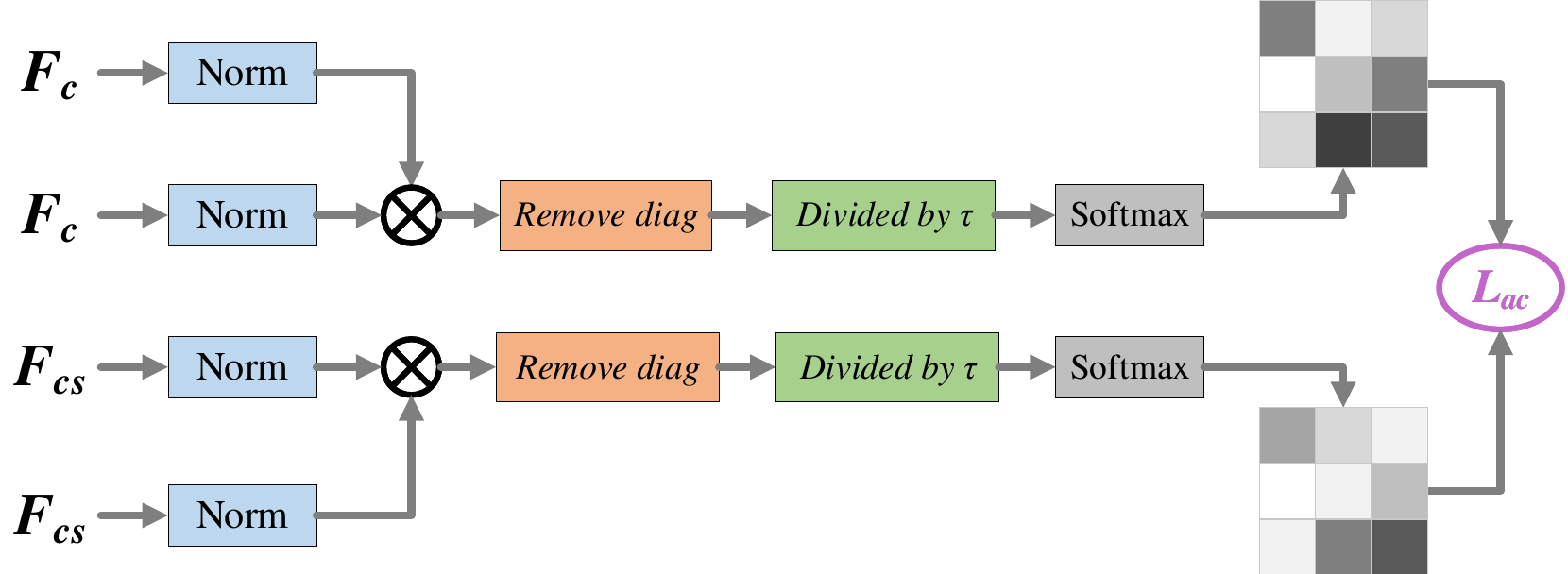}
	\caption{Illustration of our proposed attention-based content loss $\mathcal{L}_{ac}$.}
	\label{fig3}
\end{figure}

\textbf{Attention-based content loss.} $\mathcal{L}_{c}$ focuses on pulling every stylized feature point closer to the corresponding content feature point, neglecting the semantic relation among different feature points within an image. To further enhance the semantic correspondence between $I_c$ and $I_{cs}$, we first capture the semantic relation within an image based on the self-attention mechanism \cite{cheng2016long,zhang2019self} and then enforce the attention map derived from $I_{cs}$ to be consistent with that derived from $I_c$. To provide deterministic supervision signals, we use a parameter-free version of attention map without the learnable $1\times1$ convolution kernels \cite{liu2021adaattn},

\begin{equation}
A(x):=Softmax(Norm(x)^T \otimes Norm(x))
\end{equation}
where $\otimes$ denotes matrix multiplication. However, we found that the diagonal elements of the resulting attention map are very close to 1 and most of the remaining elements are 0. We argue that this is because each feature point is much more similar to itself than to other feature points and the gap is further greatly magnified by the Softmax operation. To make the attention map focus more on the inter-point relation, we remove the diagonal elements before Softmax. Nevertheless, the attention map is still a sparse matrix since the similarity between neighboring feature points is generally way above average, making other points ignored. To bridge the large gap and take more inter-point relations into account, we further scale down the absolute value of each element in the similarity matrix (before Softmax) by a factor of $\tau$ so that the resulting attention map will be a dense matrix. The above process is depicted in Figure~\ref{fig3} and formulated as,

\begin{equation}
A^{'}(x):=Softmax(\overline{diag}(Norm(x)^T \otimes Norm(x))/\tau)
\end{equation}
where $\overline{diag}$ denotes the operation of removing diagonal elements. As a result, the attention-based content loss can be defined as,

\begin{equation}
\mathcal{L}_{ac}:=\sum_{i=1}^{L} \parallel A^{'}(\phi_i(I_c)) - A^{'}(\phi_i(I_{cs})) \parallel_2
\end{equation}
For $\phi_i$, relu4\_{}1 and relu5\_{}1 layers are used in our experiments. Note that $\mathcal{L}_{ac}$ is significantly different from the content loss in STROTSS \cite{kolkin2019style}, which attempts to maintain the relative pairwise similarities between \emph{some randomly chosen locations} in an image, while the attention map used in our loss considers the semantic relations between \emph{every two different feature points} within an image and thus enables better content preservation. In addition, $\mathcal{L}_{ac}$ is based on \emph{the self-attention mechanism}, which is more effective in capturing semantic relations.

\textbf{Style loss.} The style loss is calculated by matching the mean and standard deviation of the style features to those of the stylized features,

\begin{equation}
\begin{split}
\mathcal{L}_{s}:=\sum_{i=1}^{L} \parallel \mu(\phi_i(I_{s}))-\mu(\phi_i(I_{cs})) \parallel_2 + \\
\parallel \sigma(\phi_i(I_{s}))-\sigma(\phi_i(I_{cs})) \parallel_2
\end{split}
\end{equation}
where we use relu1\_{}1, relu2\_{}1, relu3\_{}1, relu4\_{}1, and relu5\_{}1 layers to calculate this loss.

\begin{figure*}[htbp]
	\centering
	\includegraphics[width=1.0\textwidth]{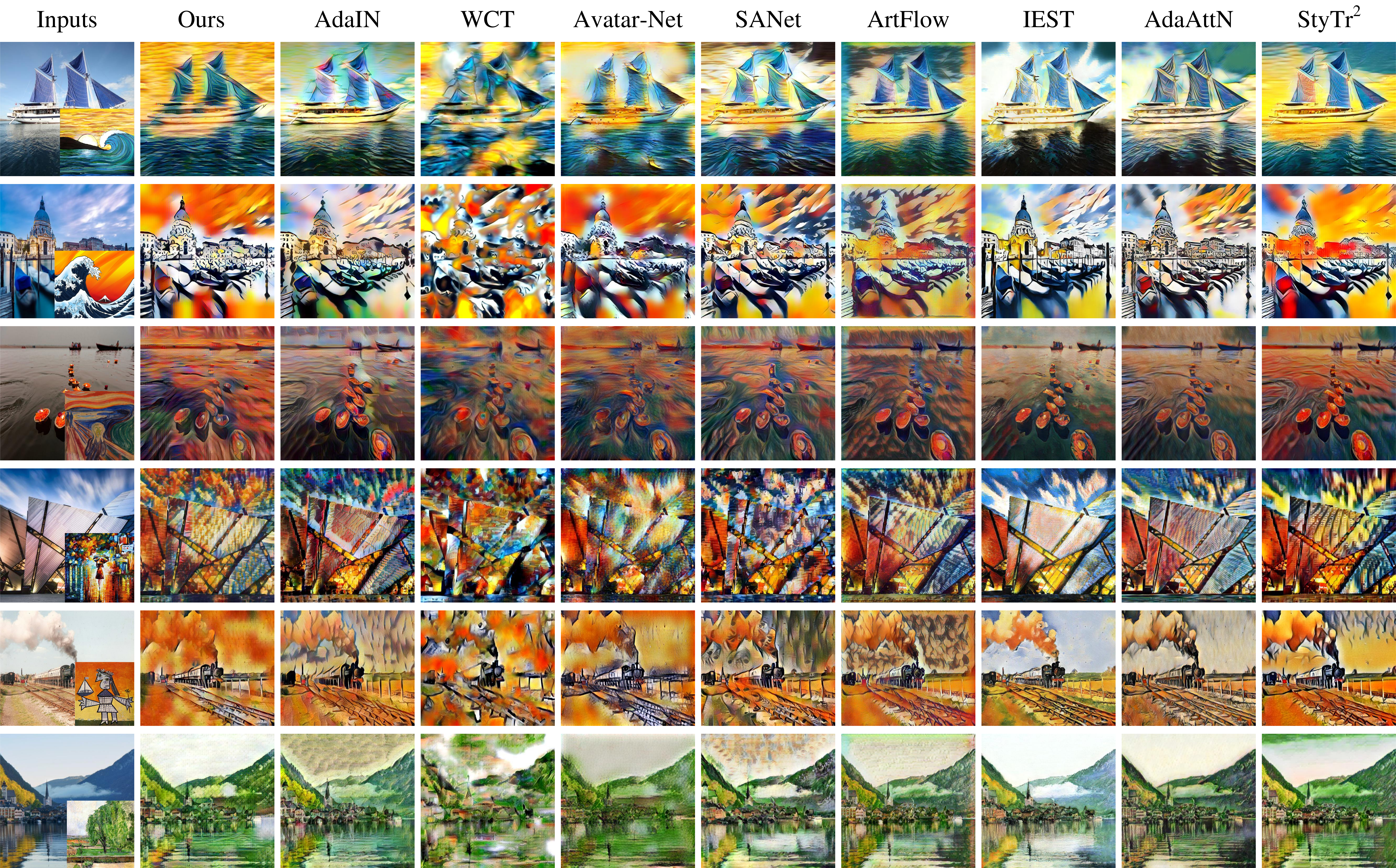}
	\caption{Qualitative comparisons. The first column shows the input content and style images. The rest of the columns show the stylization results generated with different style transfer methods. Please zoom in to compare the details.}
	\label{fig4}
\end{figure*}

\textbf{Patch-based style loss.} $\mathcal{L}_{s}$ constrains the similarity between $I_s$ and $I_{cs}$ from a global perspective. To further enhance the stylization effect from a local perspective, we propose a patch-based style loss, which encourages the style similarity between each stylized feature patch and its closest-matching style feature patch via statistics alignment,

\begin{equation}
\begin{split}
\mathcal{L}_{ps}:=\sum_{i=1}^{N} \parallel \mu(\psi_i(\phi_{r4\_1}(I_{cs})))-\mu(\psi_{NN(i)}(\phi_{r4\_1}(I_{s}))) \parallel_2 \\
+ \parallel \sigma(\psi_i(\phi_{r4\_1}(I_{cs})))-\sigma(\psi_{NN(i)}(\phi_{r4\_1}(I_{s}))) \parallel_2
\end{split}
\end{equation}
where $r4\_1$ represents $relu4\_1$, $\psi_i$ denotes the $i_{th}$ patch of the stylized features, and $\psi_{NN(i)}$ denotes its best-matching style feature patch. The patch size is the same as that in the local statistics swap operation. It is worth mentioning that the MRF loss \cite{li2016combining} also considers the local style similarity between $I_s$ and $I_{cs}$, yet it calculates the distance between two closest-matching feature patches rather than their statistics, which will inevitably introduce some semantic information from $I_s$ to $I_{cs}$.

\textbf{Identity loss.} Following \cite{park2019arbitrary,chen2021artistic,deng2022stytr2}, we also adopt an identity loss to better maintain the content structure and style characteristics simultaneously,

\begin{equation}
\begin{split}
\mathcal{L}_{identity}:=\lambda_{id1}(\parallel I_{c}-I_{cc} \parallel_2 + \parallel I_{s}-I_{ss} \parallel_2) + \\ \lambda_{id2} \sum_{i=1}^L(\parallel \phi_i(I_{c})-\phi_i(I_{cc}) \parallel_2 + \parallel \phi_i(I_{s})-\phi_i(I_{ss}) \parallel_2)
\end{split}
\end{equation}
where $I_{cc}/I_{ss}$ are the generated results when the input images are two identical content/style images. $\lambda_{id1}$ and $\lambda_{id2}$ are hyper-parameters controlling weights of their corresponding loss terms. The VGG-19 layers used here include relu1\_{}1, relu2\_{}1, relu3\_{}1, relu4\_{}1, and relu5\_{}1.

\textbf{Final objective.} We summarize all aforementioned losses and obtain the final objective of our model,

\begin{equation}
\begin{split}
\mathcal{L}:=\lambda_1 \mathcal{L}_{c} + \lambda_2 \mathcal{L}_{ac} + \lambda_3 \mathcal{L}_{s} + \lambda_4 \mathcal{L}_{ps} + \lambda_5 \mathcal{L}_{identity}
\end{split}
\end{equation}
where $\lambda_1$, $\lambda_2$, $\lambda_3$, $\lambda_4$, and $\lambda_5$ are the balancing weights for different loss terms.

\section{Experiments}

This section is organized as follows: Section~\ref{s4.1} introduces the implementation details, datasets, and baselines. Section~\ref{s4.2} and Section~\ref{s4.3} present the qualitative and quantitative results, respectively. Finally, the effect of each component in our model is explored in Section~\ref{s4.4}.

\subsection{Experimental Settings}
\label{s4.1}

\textbf{Implementation details.} As introduced in Section~\ref{s3}, our model mainly consists of three components: an encoder E, a decoder D, and a two-stage statistics-aware transformation module TSSAT. Among them, E is a fixed pre-trained VGG-19 network (up to relu4\_1) \cite{simonyan2014very}, and D is symmetrical to E. To be more specific, all pooling layers in E are replaced by nearest upsampling to form D. As for TSSAT, its architecture has been illustrated in Figure~\ref{fig2} (dashed grey box) and its input features are extracted from the $relu4\_1$ layer of VGG-19. The patch size $k$ in the local statistics swap operation is set to 5 during training, and different patch sizes can be employed to adjust local style patterns at inference. If not specifically stated, our stylization results in this paper are generated when $k$ = 5 by default. The hyper-parameter $\tau$ in Equation (8) is set to 100. The loss weights in Equation (12) and (13) are set to $\lambda_{id1}$ = 50, $\lambda_{id2}$ = 1, $\lambda_1$ = 5, $\lambda_2$ = 50000, $\lambda_3$ = 6, $\lambda_4$ = 0.5, and $\lambda_5$ = 1. We train our network using the Adam optimizer \cite{kingma2014adam} with a learning rate of 0.0001 and a batch size of 4 for 160000 iterations. Our code is available at \url{https://github.com/HalbertCH/TSSAT}.

\vspace{1.22mm}
\textbf{Datasets.} We take MS-COCO \cite{lin2014microsoft} and WikiArt \cite{karayev2013recognizing} as our content dataset and style dataset, respectively. During the training stage, we initially resize the smallest dimension of training images to 512 while maintaining the aspect ratio. Subsequently, we randomly crop patches of size 256 $\times$ 256 from these images to serve as input. In the reference stage, our method is capable of handling content and style images of any size.

\vspace{1.22mm}
\textbf{Baselines.} We select 8 state-of-the-art style transfer methods to compare with our approach, including AdaIN \cite{huang2017arbitrary}, WCT \cite{li2017universal}, Avatar-Net \cite{sheng2018avatar}, SANet \cite{park2019arbitrary}, ArtFlow \cite{an2021artflow}, IEST \cite{chen2021artistic}, AdaAttN \cite{liu2021adaattn}, and StyTr$^2$ \cite{deng2022stytr2}. For all the above baselines, we use their publicly available implementations to produce the results.

\subsection{Qualitative Results}
\label{s4.2}

We present qualitative stylization results of different style transfer methods in Figure~\ref{fig4} for comparison. It can be observed that AdaIN \cite{huang2017arbitrary} often captures insufficient style patterns and introduces some abrupt colors that do not exist in the style image (\emph{e.g.}, $1^{st}$, $2^{nd}$, and $6^{th}$ rows). WCT \cite{li2017universal} has severe problems with content preservation (\emph{e.g.}, $2^{nd}$, $4^{th}$, and $6^{th}$ rows). Avatar-Net \cite{sheng2018avatar} suffers from the content structure blur and style pattern distortion issues (\emph{e.g.}, $1^{st}$, $2^{nd}$, and $4^{th}$ rows). SANet \cite{park2019arbitrary} sometimes introduces undesired semantic structures from the style image to the stylization result (\emph{e.g.}, $3^{rd}$, $4^{th}$, and $5^{th}$ rows). ArtFlow \cite{an2021artflow} tends to produce unwanted artifacts in relatively smooth regions (\emph{e.g.}, $2^{nd}$, $3^{rd}$, and $5^{th}$ rows). The results of IEST \cite{chen2021artistic} are generally less stylized with limited colors and textures (\emph{e.g.}, $2^{nd}$, $4^{th}$, and $5^{th}$ rows). For AdaAttN \cite{liu2021adaattn}, there is an obvious style deviation between the style image and the stylized image generated by it (\emph{e.g.}, $1^{st}$, $5^{th}$, and $6^{th}$ rows). The results of StyTr$^2$ \cite{deng2022stytr2} usually have the problem of color oversaturation, resulting in inconsistent colors with the style image (\emph{e.g.}, $1^{st}$, $2^{nd}$, and $3^{rd}$ rows). In comparison, our method TSSAT not only captures accurate and adequate style patterns, but also retains clear and clean content structures, as shown in the $2^{nd}$ column of Figure~\ref{fig4}. \emph{\textbf{Please zoom in to compare the details.}}

\begin{figure}[t]
	\centering
	\includegraphics[width=1.0\columnwidth]{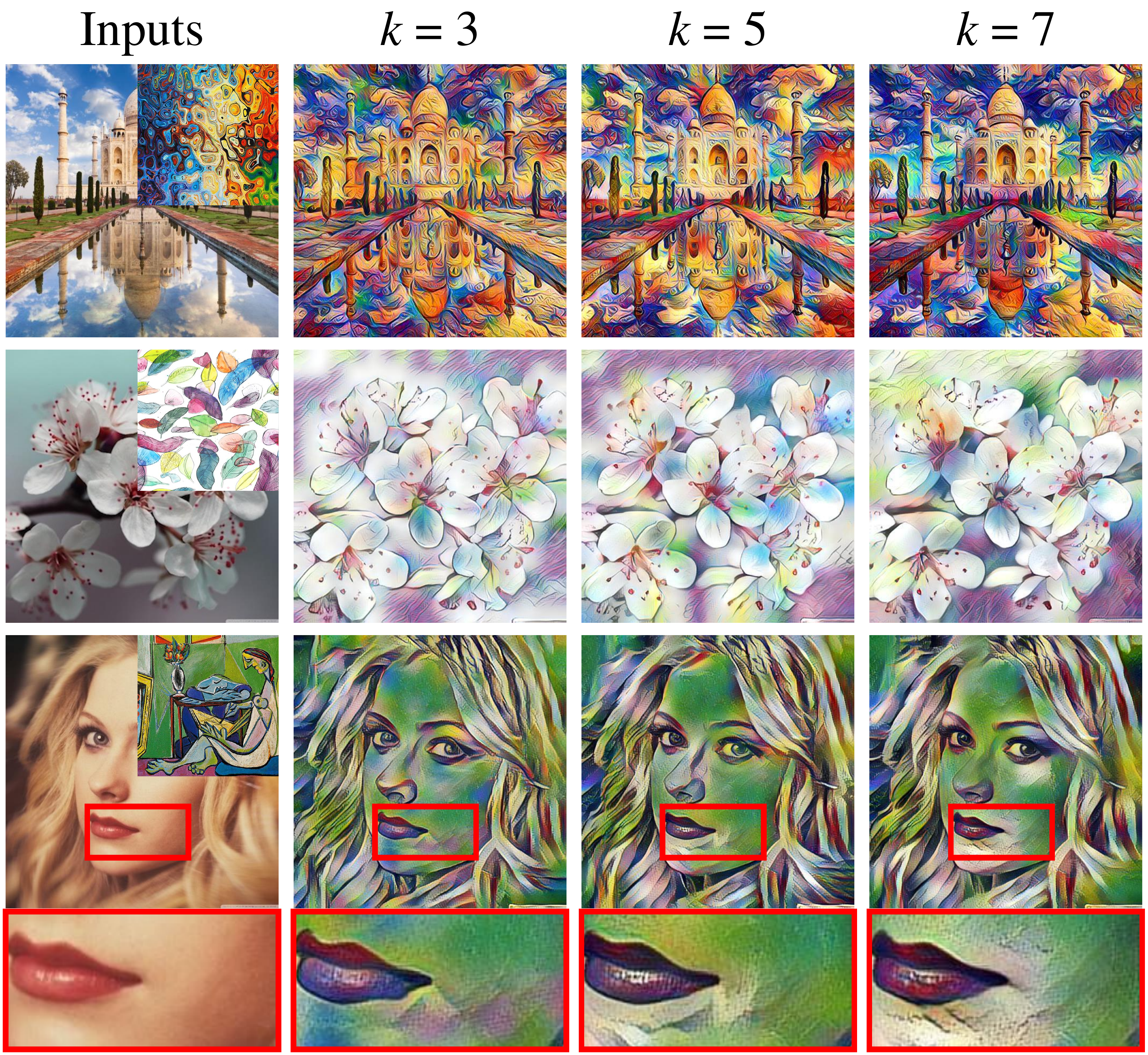}
	\caption{Stylization results with different patch sizes $k$.}
	\label{fig5}
\end{figure}

As introduced in Section~\ref{s3.2}, our proposed TSSAT also allows flexible style pattern modulation by adjusting the patch size $k$ in the local statistics swap stage. Note that our model can adapt to different patch sizes at inference once trained with one patch size. Therefore, it is very convenient and efficient for our model to produce diverse stylization results with different patch sizes, as shown in Figure~\ref{fig5}. We can see that the stylized images are more colorful and vivid when the patch size is small and become cleaner and neater when the patch size is bigger. The zoom-in regions demonstrate the change of local style patterns more clearly.

\subsection{Quantitative Results}
\label{s4.3}

The qualitative results presented above could be subjective. In this section, we adopt several quantitative metrics to conduct more comprehensive and objective evaluations.

\textbf{Perceptual distance and GELP.} Perceptual distance \cite{gatys2016image,johnson2016perceptual} estimates the multi-level feature distances between the content and stylized images. It is usually taken as the content loss by existing style transfer methods, including our and competing methods. Following \cite{wu2020efanet,an2021artflow,deng2022stytr2}, here we adopt it to measure the performance of content preservation. Meanwhile, to measure the performance of style transformation, we employ the GELP metric \cite{wang2021evaluate,wang2022aesust} that takes both global style effects (including global colors and holistic textures) and local style patterns (including the similarity and diversity of the local style patterns) into consideration. We randomly select 50 content-style pairs for each method and report the average perceptual distance and GELP score in Table~\ref{table1}. As we can see, our proposed TSSAT obtains the lowest perceptual distance and the third-highest GELP score. The results indicate that our method achieves the best trade-off between content preservation and style transformation, which is consistent with the visual comparisons in Figure~\ref{fig4}. We also compare the performance of our method under different patch sizes $k$. It is easy to find that: the bigger the patch size, the better the content preservation; the smaller the patch size, the better the style transformation.

\textbf{Preference score.} We further perform a user study \cite{park2019arbitrary,chen2020creative,an2021artflow,chandran2021adaptive,zhang2022exact,yang2022industrial,wang2023microast} to investigate user preference over different stylization results. Specifically, we first choose 15 content images and 10 style images to form 150 content-style pairs. Then, we randomly sample 20 content-style pairs for each subject and synthesize 9 different stylized images for each pair using 9 style transfer methods (including our method and 8 baselines). Next, we ask the subject to indicate his/her favorite stylization result for each content-style pair. Finally, we collect 1000 votes from 50 subjects and show the percentage of votes for each method in Table~\ref{table1}, where we can observe that our method achieves preferable performance than competitors by a significant margin. 

\textbf{Efficiency analyses.} We compare the efficiency of our method with prior works in the bottom row of Table~\ref{table1}. All the methods are tested on a single Nvidia GeForce RTX 3090 GPU with the image size of 512$\times$512. It can be observed that the speed of our method is comparable with the state-of-the-art methods such as ArtFlow \cite{an2021artflow} and StyTr$^2$ \cite{deng2022stytr2}. Moreover, the speed of our method can be further accelerated by increasing the patch size $k$.

\begin{table*}[htbp]
	\caption{Quantitative comparisons. Dis. stands for distance. We show the best results in \textbf{bold}, the second-best results with a star*, and the third-best results with an \underline{underline}.}
	\centering
	\renewcommand{\arraystretch}{1.29}
	\resizebox{1.0\textwidth}{!}{
		\begin{tabular}{c|c|c|c|c|c|c|c|c|c|c|c}
			\hline
			
			\multirow{2}{*}{}    & \multirow{2}{*}{AdaIN}    & \multirow{2}{*}{WCT}    & \multirow{2}{*}{Avatar-Net}    & \multirow{2}{*}{SANet}    & \multirow{2}{*}{ArtFlow}    & \multirow{2}{*}{IEST}    & \multirow{2}{*}{AdaAttN}     & \multirow{2}{*}{StyTr$^2$}
			& \multicolumn{3}{c}{Ours} \\ \cline{10-12}
			& & & & & & & & & $k$ = 3   & $k$ = 5    & $k$ = 7 \\
			
			\hline
			Perceptual Dis. $\downarrow$  & 2.061             & 2.828          & 2.297             & 2.251      & 2.052         & 1.876*    & 2.191      & 1.958            & 2.019   & \underline{1.880}    & \textbf{1.778}  \\
			GELP Score $\uparrow$         & 1.446             & 1.457          & 1.488*    & \textbf{1.511}      & 1.451         & 1.382    & 1.438      & 1.461    & \underline{1.481}   & 1.476    & 1.469  \\
			Preference (\%) $\uparrow$    & 0.059             & 0.053          & 0.061             & 0.094      & 0.070         & \underline{0.152}    & 0.105      & 0.174*            & -   & \textbf{0.232}        & -      \\
			\hline
			Time (sec) $\downarrow$  & \textbf{0.062}             & 0.997          & 0.308             & \underline{0.077}      & 0.341         & 0.074*    & 0.112      & 0.401            & 0.484   & 0.337    & 0.329  \\
			\hline
		\end{tabular}
	}
	\label{table1}
\end{table*}

\subsection{Ablation Study}
\label{s4.4}

\begin{figure*}[htbp]
	\centering
	\includegraphics[width=1.0\textwidth]{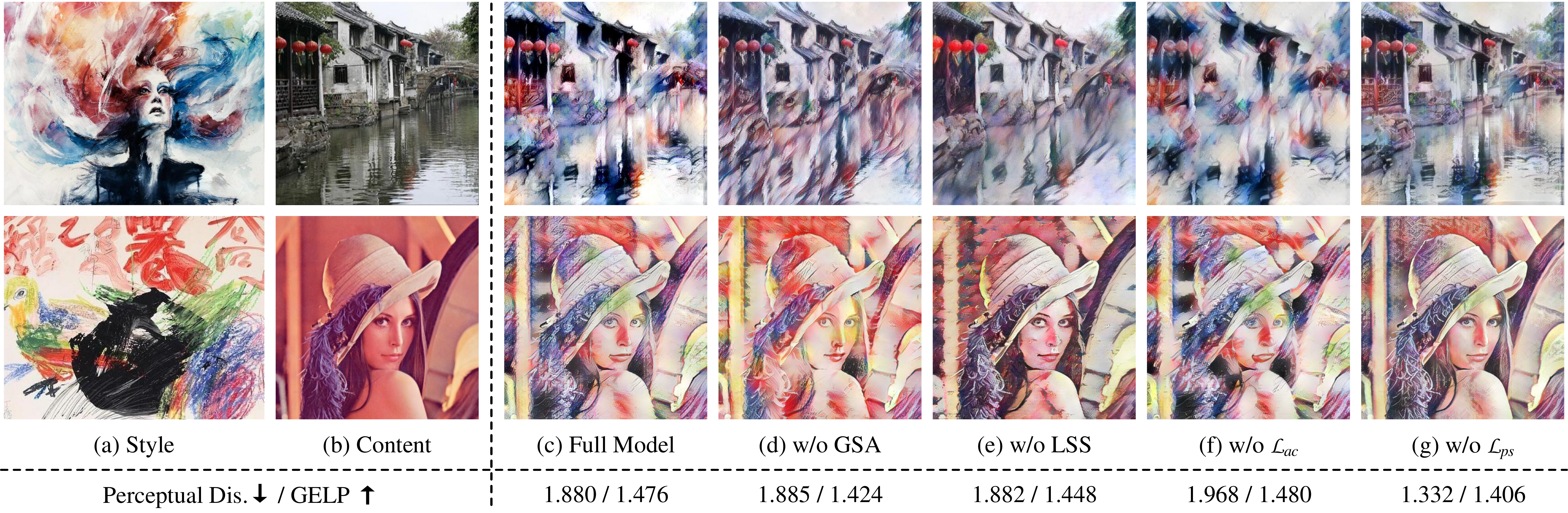}
	\caption{Ablation study results. The first two columns show the style and content images, respectively. The rest columns show the stylization results generated by our model under different settings.}
	\label{fig6}
\end{figure*}

\textbf{Analyses of the TSSAT module.} As introduced in Section~\ref{s3.2}, the TSSAT module consists of a global statistics alignment ($abbr.$ GSA) stage and a local statistics swap ($abbr.$ LSS) stage. To study their effects, we compare the stylization results of our method with and without GSA/LSS in Figure~\ref{fig6} (c-e). We can see that without GSA, the model neglects the global style distribution of the style image. Without LSS, the model fails to capture abundant and fine-grained local colors and texture patterns. This is because GSA and LSS are responsible for global style learning and local style capturing, respectively. We can also see that the LSS stage performed after GSA influences the global style only to a restricted extent and will not result in global style deviation. We can get an explanation from the relation between the two stages: GSA builds the global style foundation and LSS just further enriches local style details based on the foundation. Above analyses are also supported by the quantitative results reported in the last row.


\textbf{Loss analyses.} To investigate the influence of the attention-base content loss $\mathcal{L}_{ac}$ and the patch-based style loss $\mathcal{L}_{ps}$, we remove them from our model and show the experimental results in Figure~\ref{fig6} (f) and (g). It can be observed that without $\mathcal{L}_{ac}$, the content structures of the stylized image become less clear, and notable distortions are introduced. The results demonstrate the importance of $\mathcal{L}_{ac}$ in content preservation. In addition, we also find that without $\mathcal{L}_{ps}$, the stylization results become less colorful and vivid, and lots of local style information is lost. It indicates that $\mathcal{L}_{ps}$ is of great significance in local style learning. The qualitative results, together with the quantitative results reported in the last row, verify that only the full model can achieve satisfying performance in both content preservation and style transformation.

\section{Conclusion and Limitation}
In this paper, we propose a Two-Stage Statistics-Aware Transformation (TSSAT) module and two loss functions to improve the style transformation and content preservation effect of artistic style transfer. The contribution of TSSAT is the idea of harnessing feature statistics to first build the global style foundation (in the global statistics alignment stage) and then further enrich local style details (in the local statistics swap stage), simulating the drawing process of humans. The feature statistics we adopt in this paper are mean and variance, and more alternatives can be explored in the future. The attention-based content loss enables better content preservation by enforcing the semantic relation in the content image to be retained during stylization. The patch-based style loss facilitates local style learning by encouraging the similarity between each stylized feature patch and its closest-matching style feature patch. Extensive experiments demonstrate the effectiveness and superiority of our proposed method. 

A main limitation of this work is that the proposed method is not fast enough to achieve real-time style transfer. This is because the local statistics swap operation in our TSSAT module needs to be conducted for many times between different feature patches, which is kind of time consuming (the smaller the patch size, the slower the speed). We will take the efficiency issue as our future work and try to simplify the local statistics swap operation for higher execution speed.


\begin{acks}
This work was partially supported by the National Science Foundation of China (Grant Nos. 62072242).
\end{acks}

\bibliographystyle{ACM-Reference-Format}
\balance
\bibliography{acmart}

\end{document}